\documentclass{llncs} %
\usepackage{graphicx}
\usepackage{amsmath,amssymb} % define this before the line numbering.
\usepackage{color}
\usepackage[width=122mm,left=12mm,paperwidth=146mm,height=193mm,top=12mm,paperheight=217mm]{geometry}

\usepackage{times}
\usepackage{epsfig}
\usepackage{epstopdf}
\usepackage{multirow}
\usepackage{color}

\usepackage{url}
\urldef{\mailsa}\path|{qingshanxu, wenbingtao}@hust.edu.cn|

\begin{document}
% \linenumbers
\pagestyle{headings}
\mainmatter

\title{Multi-View Stereo with Asymmetric Checkerboard Propagation and Multi-Hypothesis Joint View Selection} % Replace with your title

%\titlerunning{ECCV-18 submission ID \ECCV18SubNumber} %
%
%\authorrunning{ECCV-18 submission ID \ECCV18SubNumber} %

%\author{Anonymous ECCV submission}
\author{Qingshan Xu, Wenbing Tao}
%\institute{Paper ID \ECCV18SubNumber} %
\institute{National Key Laboratory of Science and Technology on Multi-spectral Information Processing, School of Automation, Huazhong University of Science \& Technology \\
\mailsa}

\maketitle

\begin{abstract}
In computer vision domain, how to fast and accurately perform multi-view stereo (MVS) is still a challenging problem. In this paper we present a fast yet accurate method for 3D dense reconstruction, called AMHMVS, built on the PatchMatch based stereo algorithm. Different from the regular symmetric propagation scheme, our approach adopts an asymmetric checkerboard propagation strategy, which can adaptively make effective hypotheses expand further according to the confidence of current neighbor hypotheses. In order to aggregate visual information from multiple images better, we propose the multi-hypothesis joint view selection for each pixel, which leverages a cost matrix based on the multiple propagated hypotheses to robustly infer an appropriate aggregation subset parallel. Combined with the above two steps, our approach not only has the capacity of massively parallel computation, but also obtains high accuracy and completeness. Experiments on extensive datasets show that our method achieves more accurate and robust results, and runs faster than the competing methods.
%\dots
\end{abstract}

\section{Introduction}

Multi-view stereo (MVS) has traditionally been a topic of interest in computer vision for decades. It can be defined as establishing dense correspondence between multiple calibrated images, which results in a dense 3D reconstruction. Over the last few years, much effort has been put into improving the the quality of dense 3D reconstruction~\cite{Alpher01,Alpher02,Alpher03,Alpher04,Alpher05,Alpher06,Alpher07,Alpher08,Alpher09,Alpher10,Alpher11,Alpher12} and some work has achieved impressive visual effect. However, considering the challenge of large-scale data, heterogeneous scene illuminations, different view points, variable scene contents and image registration errors, how to fast and accurately perform multi-view stereo in computer vision domain still remains an open problem.

{\em PatchMatch}~\cite{Alpher18}, whose main idea is to randomly initialize a correspondence field and then iteratively propagate the good correspondence between neighbors, was first introduced into stereo matching by Bleyer {\em et al.}~\cite{Alpher13} to efficiently find a good 3D plane at each pixel. Then many extending methods~\cite{Alpher08,Alpher09,Alpher10} have emerged in multi-view stereo for improving computation efficiency and reconstruction accuracy. The studies have demonstrated that, the propagation scheme and pixelwise view selection is a core component in fast and accurate {\em PatchMatch} based multi-view stereo methods. To obtain an accurate dense 3D model, Zheng {\em et al.}~\cite{Alpher08} and Sch{\"o}nberger {\em et al.}~\cite{Alpher10} skillfully design the probabilistic graphical model to infer the aggregation subset of the source image for every particular pixel in the reference image. Considering their propagation scheme is sequential in nature, Galliani {\em et al.}~\cite{Alpher09} propose a diffusion-like propagation to efficiently process a large amount of correspondence between images, which can make full use of the powerful parallel computation of GPU. However, their method fails to accurately infer the view selection subset at each pixel and decreases the reconstruction accuracy. Because of the difficulty in dealing with the delicate pixelwise view selection and massively parallel propagation at the same time, these methods are still struggling in taking both efficiency and accuracy into account.

In this work, we present a multi-view stereo method combining the asymmetric checkerboard propagation with the multi-hypothesis joint view selection to deliver fast, accurate, and dense 3D reconstructions. A key observation is that, with the evolution of iterative asymmetric checkerboard propagation, the costs of true hypotheses will appear to be much lower and stabler. To get the more effective candidate hypotheses, we first introduce the asymmetric diffusion-like propagation to make true hypotheses expand further. After that, we construct a cost matrix based on these current deemed effective hypotheses to infer a good aggregation subset, and then pick the true hypotheses according to the temporary selected views. Combined with the above two strategies, our approach utilizes the parallelism of GPU better and gets more accurate depth estimation for every pixel.

The contributions of this paper include (1) We propose the asymmetric checkerboard propagation. We show that it is more effective to expand reasonable hypotheses further. (2) We propose the multi-hypothesis joint view selection to progressively infer a aggregation subset at each pixel parallel based on the correlation among true hypotheses and views. It avoids sequential probabilistic graphical model inference and is more suited to be combined with the above propagation scheme. (3) We show the effectiveness and efficiency of our stereo approach by achieving high accuracy but running faster than other competing methods on extensive experiments.

\section{Related Work}

Inspired by the comprehensive benchmark datasets and evaluation methods~\cite{Alpher14,Alpher15,Alpher16,Alpher31}, a lot of research has been focusing on multi-view stereo. Here, we do not review the entire literature on it and only discuss the related approaches. We suggest readers referring to publications like~\cite{Alpher14,Alpher35} for an overview of multi-view stereo methods.

Multi-view stereo deals with depth estimation from multiple calibrated images according to visual similarity. Generally speaking, there are four important stages in multi-view stereo methods, including view selection, propagation scheme, patch matching and depth map fusion. As a key component in multi-view stereo, view selection chooses the individual view aggregation subset for each pixel. In ~\cite{Alpher17}, Kang {\em et al.} proposed to include only the best 50\% of all {\em N} cost values and Galliani {\em et al.}~\cite{Alpher09} slightly changed this to a fixed parameter {\em K} to generate the view subset. Zheng {\em et al.}~\cite{Alpher08} showed in their experiments that, no matter how to adjust the parameter {\em K} can not achieve satisfactory results. Goesele {\em et al.}~\cite{Alpher02} first constructed the global view sets according to the geometry of points with known depths, which called Ground Control Points (GCPs). Then they further chose the local view subsets in term of the Normalized Cross Correlation (NCC) and the epipolar geometry constraint. It seriously depended on the GCPs and is not suitable to estimate depth parallel. Zheng {\em et al.}~\cite{Alpher08} proposed to construct the probabilistic graphical model to infer the appropriate aggregation subset combined with the variational inference. Built on this work, Sch{\"o}nberger {\em et al.}~\cite{Alpher10} introduced the normal estimation to the probabilistic graphical model and simultaneously considered a variety of photometric and geometric priors.

As for the propagation scheme, the seed-and-expand based methods~\cite{Alpher01,Alpher02} expanded the reliable seeds to their neighbors to generate new seeds. Although these work avoided processing textureless regions and saved some computations, the expanding scheme is irregular with the number and position of existing seeds. Thus, they can hardly make full use of parallelism computing power of GPU. Unlike the seed-and-expand scheme, the {\em PatchMatch} based stereo methods can make better use of parallel computing ability of GPU. Bailer {\em et al.}~\cite{Alpher04} adopted the top left to bottom right propagation while Bleyer {\em et al.}~\cite{Alpher13} used the rightward propagation. In order to meet the requirement of progressive inference progress, Zheng {\em et al.}~\cite{Alpher08} and Sch{\"o}nberger {\em et al.}~\cite{Alpher10} alternatively performed upward/downward propagations during odd iterations and perform rightward/leftward propagations during even propagations. As pointed out in ~\cite{Alpher09}, these standard {\em PatchMatch} propagation is sequential because every point is dependent on the previous one. Instead, they proposed the diffusion-like scheme in a checkerboard pattern specifically tailored to multi-core architectures such as GPU processors. Unfortunately, the reasonable hypotheses can only be spread to a limited distance in this scheme.

The seminal work, {\em PatchMatch}, was first proposed by Barnes {\em et al.}~\cite{Alpher18} and then introduced into stereo matching by Bleyer {\em et al.}~\cite{Alpher13}. They showed that PatchMatch has high computational efficiency. Hereafter, many multi-view stereo methods~\cite{Alpher08,Alpher09,Alpher10} are rapidly developed based on the idea. Galliani {\em et al.}~\cite{Alpher09} first proposed to run PatchMatch stereo in scene space for multi-view stereo. They randomly initialized 3D scene planes to avoid epipolar rectification and allow the data cost to directly aggregate evidence from multiple views. Then they verified good hypotheses and propagated them to their neighbors iteratively. Zheng {\em et al.}~\cite{Alpher08} also employed the PatchMatch algorithm to jointly optimize the view selection and depth estimation. Without considering normal, their method ignored the existence of slanted surfaces and used the fronto-parallel scene structure assumption, which led to artifacts for oblique structures~\cite{Alpher22}. Besides depth estimation, Sch{\"o}nberger {\em et al.}~\cite{Alpher10} simultaneously considered the normal estimation like ~\cite{Alpher09}, which meant that they took slanted surfaces into account. Then both were optimized through propagating and sampling.

Depth map fusion merges the individual depth map of each calibrated view into a single point cloud and mitigates the wrong depth values among the individual estimates. Jancoseck and Pajdla~\cite{Alpher20} computed depth maps by implementing the plane-sweeping approach and construct the s-t graph to fuse depth maps into a surface. Shen~\cite{Alpher06} fused the depth maps by neighboring depth map test. For each pixel in current camera, he back projected it to 3D point and reprojected the 3D point to 2D image coordinate. Then he rejected the redundancies and outliers according to the consistency check. Hu and Mordohai~\cite{Alpher19} first initialized depth and confidence estimation and did depth map fusion by explicitly modeling geometric and correspondence uncertainty. Galliani {\em et al.}~\cite{Alpher09} generated the best possible individual depth maps, and then merged them into a complete point cloud in a straightforward manner, which leveraged the depth and normal estimation to implement consistency check. Zach~\cite{Alpher21} employed the variational formulation for the surface reconstruction task and solved it by parallelized gradient descent method on the GPU. Sch{\"o}nberger {\em et al.}~\cite{Alpher10} proposed the graph-based filtering and fusion of depth and normal maps. Their fusion method initialized a new cluster by using the node with maximum photometric and geometric support, and recursively collected connected nodes that satisfied the constraints. Then they got different clusters of consistent pixels in the graph, and fused the cluster's element by median processing.

\begin{figure*}[t]
\begin{center}
%\fbox{\rule{0pt}{2in} \rule{.9\linewidth}{0pt}}
   \includegraphics[width=0.98\linewidth]{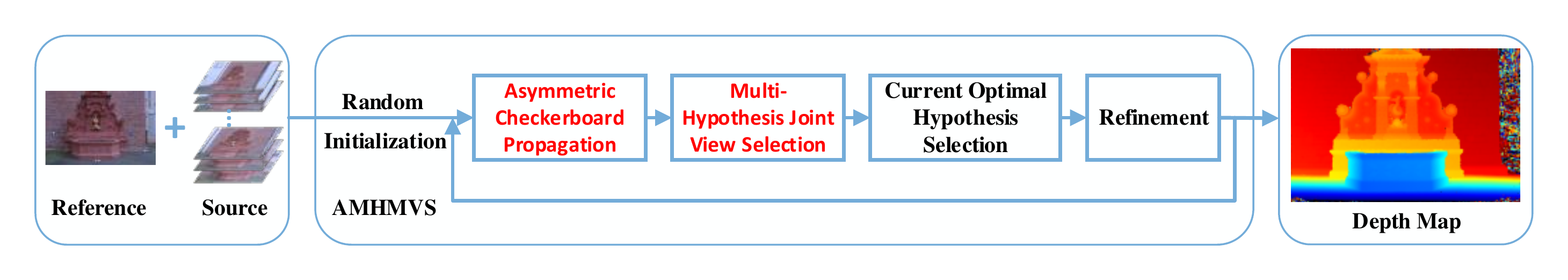}
\end{center}
   \caption{Overview of our proposed stereo method, which is utilized to generate the depth map for every reference image in turn. Our key strategies are highlighted in red.}
\label{fig:overview}
\end{figure*}

\section{AMHMVS Algorithm}

In this section, we present our multi-view stereo method and discuss its main features. Our multi-view stereo method builds upon the {\em Gipuma}~\cite{Alpher09} stereo framework and works with new propagation and view selection scheme. An overview of our proposed stereo method is given in Fig.~\ref{fig:overview}. We first describe the confidence-guided asymmetric checkerboard propagation in Section ~\ref{ACP}, and then detail the multi-hypothesis joint view selection of our approach in Section ~\ref{MHJ}.

\subsection{Asymmetric Checkerboard Propagation}\label{ACP}

\begin{figure}[t]
\begin{center}
   %\includegraphics[width=0.8\linewidth]{egfigure.eps}
   \includegraphics[width=0.24\linewidth]{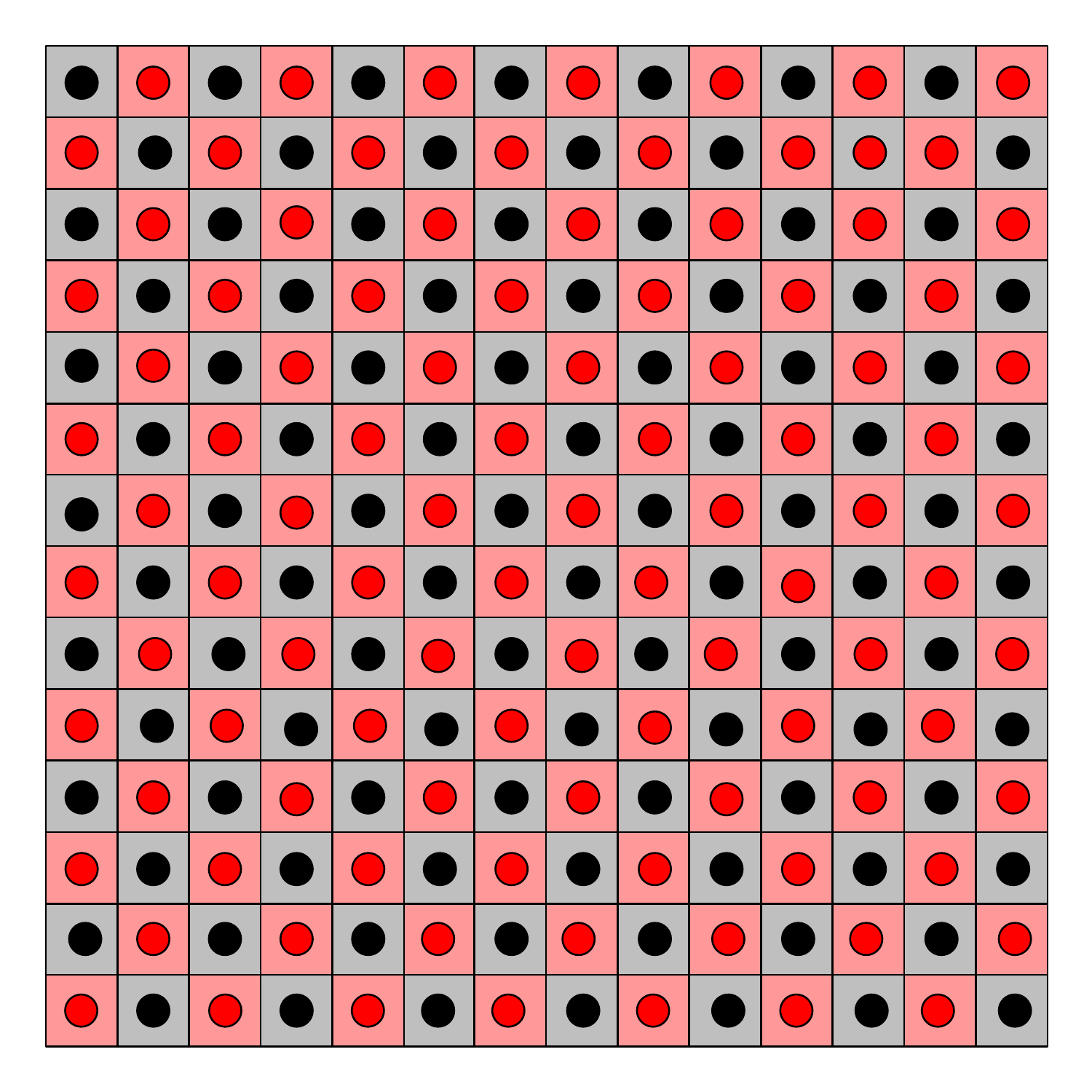}
   \includegraphics[width=0.24\linewidth]{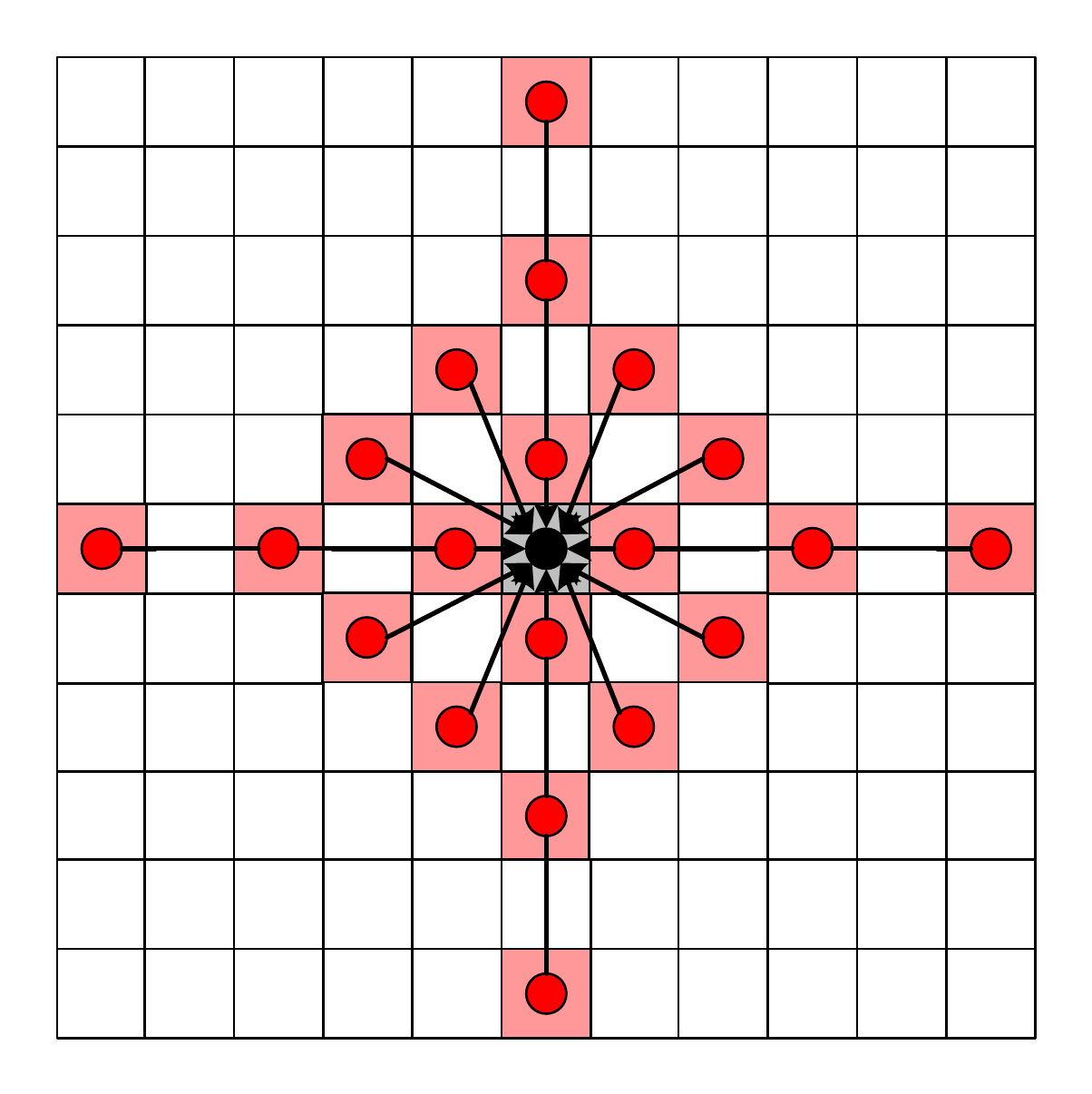}
   \includegraphics[width=0.24\linewidth]{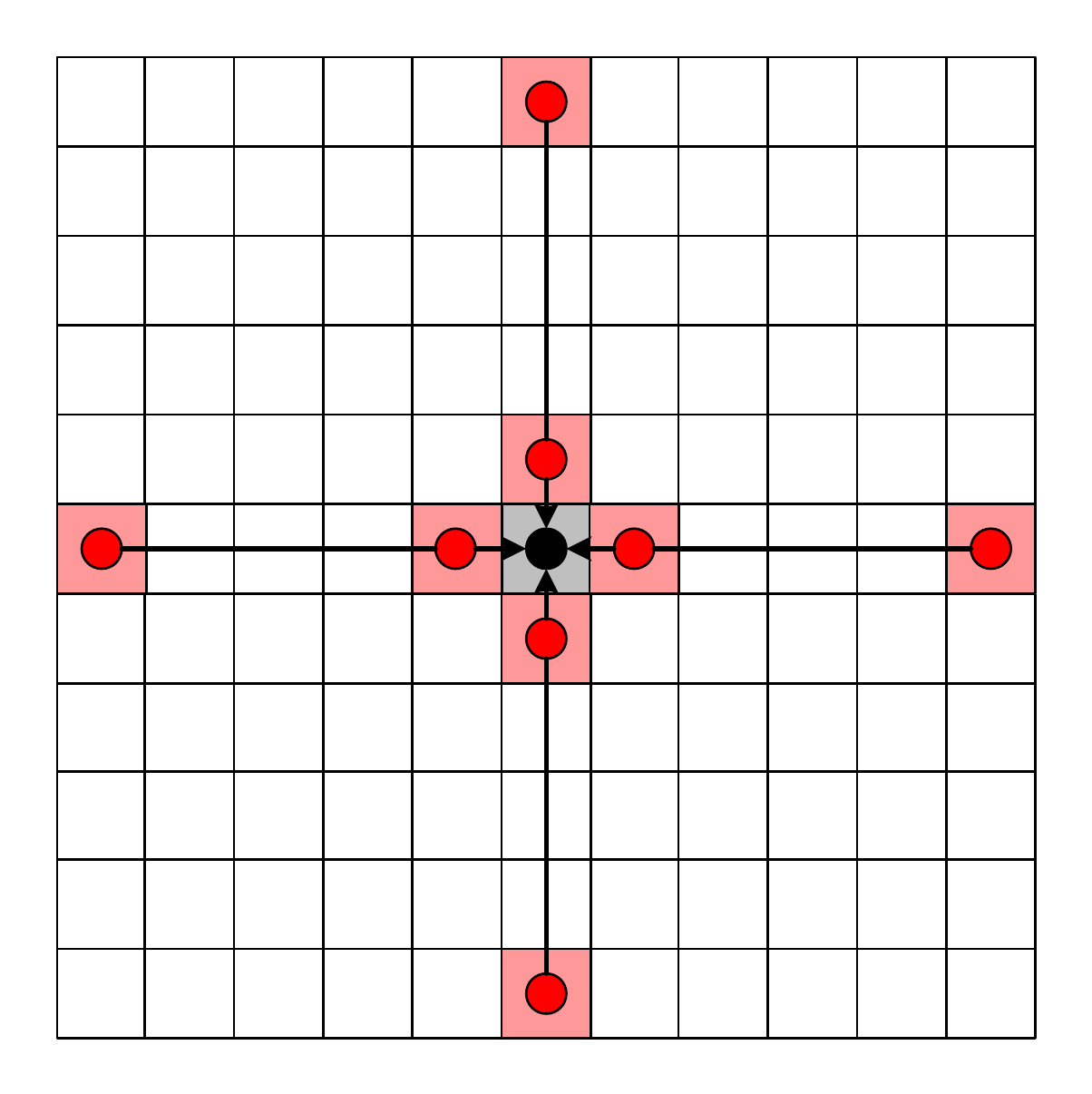}
   \includegraphics[width=0.235\linewidth]{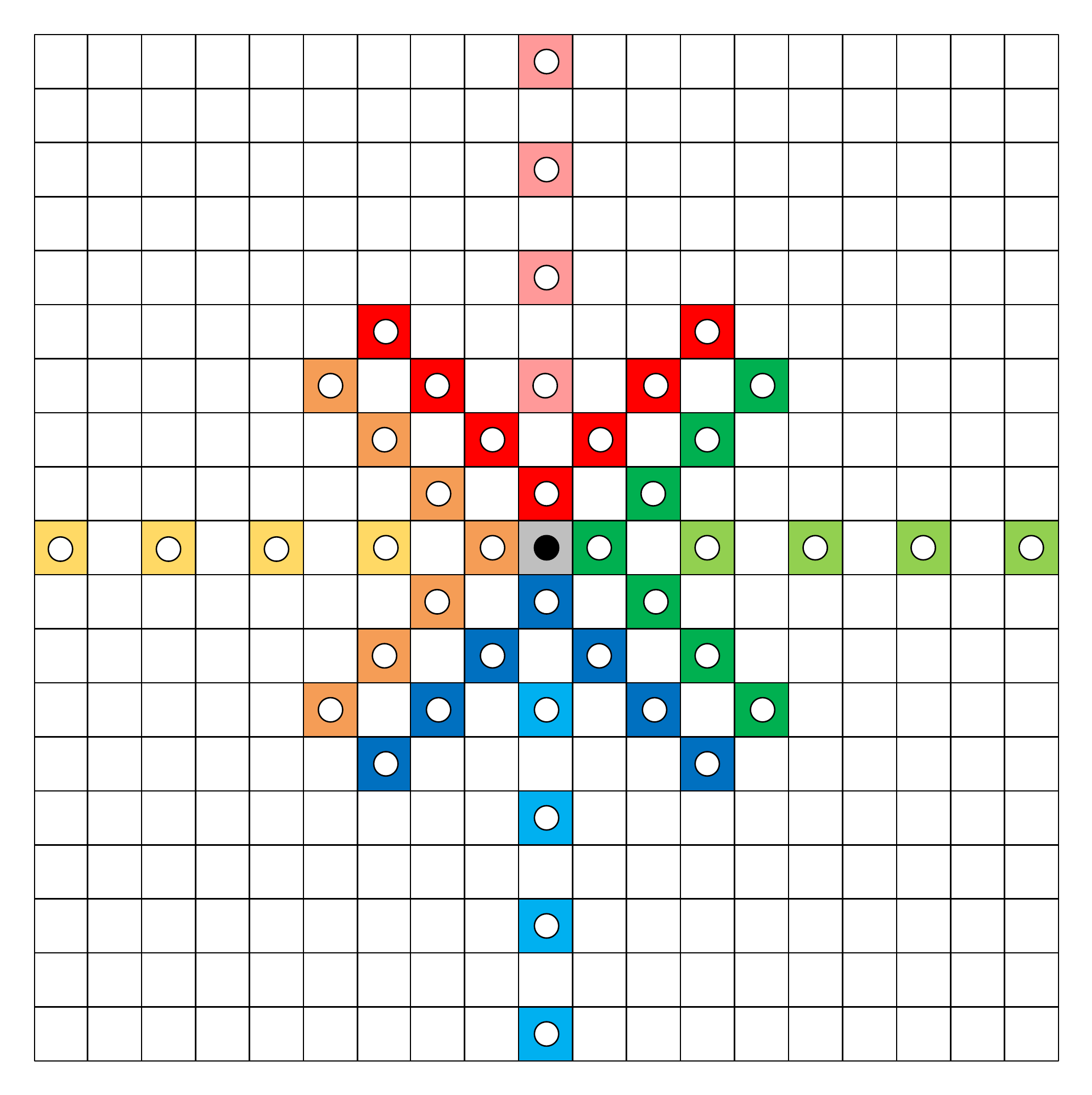}
   \qquad\qquad\qquad\qquad\textrm{(a)}\qquad\qquad\qquad\qquad\quad\textrm{(b)}\qquad\qquad\qquad\quad\quad\textrm{(c)}\qquad\qquad\qquad\quad\quad\textrm{(d)}\qquad\qquad
\end{center}
   \caption{The propagation scheme: (a) The red-black checkerboard pattern for updating the depth and normal of black pixels using the red pixels and vice versa.~\cite{Alpher09} (b) The standard version of checkerboard diffusion-like propagation.~\cite{Alpher09} (c) The fast version of checkerboard diffusion-like propagation.~\cite{Alpher09} (d) Our proposed asymmetric checkerboard propagation (Best viewed in color).}
\label{fig:ACP}
\label{fig:onecol}
\end{figure}

The standard {\em PatchMatch} propagation scheme propagates information diagonally cross the image, and alternates between a pass from top left to bottom right and a pass in the opposite direction. As shown in Fig.~\ref{fig:ACP}(a), (b) and (c), distinguished from the standard {\em PatchMatch} propagation scheme, {\em Gipuma} stereo method exploits the diffusion-like scheme to do message-passing. It considers all pixels of the reference image as the red-black grids of the checkerboard. Then it updates all black ones by leveraging the individual hypotheses from their local regular red neighbors. The all red ones are updated in a similar way. It avoids the sequential nature of other parallel propagation scheme~\cite{Alpher04,Alpher08,Alpher10,Alpher13} by the checkerboard propagation pattern.

%figure 1

Although {\em Gipuma} stereo proposes two versions of the diffusion-like checkerboard propagation including a standard one and a fast one, it can still only propagate the good hypotheses to a limited distance. Inspired by the work of Sun {\em et al.}~\cite{Alpher28,Alpher29}, we propose the confidence-guided asymmetric checkerboard propagation. The message passing is asymmetric and adaptive for Belief Propagation in stereo matching. Similarly, the message in the framework of probabilistic graphical model~\cite{Alpher08,Alpher10} can be expanded as far as possible in the depth approximately continuous regions. In contrast, the message passing will be stopped in the discontinuity regions. As the checkerboard propagation is regular and symmetric, it suffers from the problem of limited propagation.

We hope to propagate good hypotheses further but still keep the checkerboard pattern that is important for GPU parallel computation. Thus, we divide the neighbor areas of each pixel into two parts including closer region (dark color) and distant region (light color) as depicted in Fig.~\ref{fig:ACP}(d). The reason for this is that the hypotheses in the closer region represent the possibility of depth continuity while the hypotheses in the distant region denote the possibility of depth mutation. Moreover, in order to maintain the efficiency, we only propagate 8 hypotheses, corresponding to 8 regions of different colors.

Then we sample the most possible hypothesis from the individual region according their current matching costs. The smaller the matching cost of a hypothesis, the more likely it is to be a true hypothesis. In fact, if good hypotheses can expand further in the regular and symmetric diffusion-like checkerboard propagation scheme, we can give a chance to these hypotheses to propagate further earlier. Consequently, many wrong hypothesis can stop propagating earlier, and it will leave a opportunity to more hypothesis candidates to have a try. Our propagation scheme to some extent simulates the above process. This will make the message-passing become asymmetric and adaptive, leading to propagate good hypotheses as far as possible and generate more hypotheses. In practice, we sample the closer hypotheses from 7 hypothesis candidates, and the distant hypotheses from 11 hypothesis candidates in the extend upward, downward, leftward and rightward directions.

\subsection{Multi-Hypothesis Joint View Selection}\label{MHJ}

For multi-view stereo, a key component is how to best aggregate multiple pairwise photometric information into a unified cost representation. A common way to handle the aggregation of multiple views is to heuristically select views with a much smaller cost, and take the mean of their costs as the unified cost representation. This method leads to following phenomenon: (\romannumeral1) the selected view of every propagated hypothesis is different, (\romannumeral2) the mean processing makes the selected views have the same importance, (\romannumeral3) it seriously depends on the predefined parameter ${K}$. These phenomenon result in the corresponding drawbacks: (\romannumeral1) the selected best hypothesis is biased by the different comparison baseline every time, (\romannumeral2) some good hypotheses are always harmed by the bad ones, (\romannumeral3) it can not truncate ${K}$ adaptively. To alleviate the above problems and keep the checkerboard propagation, we propose a multi-hypothesis joint view selection scheme. It first constructs a cost matrix based on the above selected hypotheses to infer the good aggregation view subset, and then pick true hypotheses according to the temporary selected views.

Given a set of source images ${\boldsymbol{\mathcal{X}}}^{src}=\{\mathcal{X}^{z}|z=1,\ldots,Z\}$ with known calibrated camera parameters. The goal of our method is to estimate the depth map of the reference image ${{\mathcal{X}}}^{ref}$. We first compute matching costs at pixel ${\bf x}$ over all the source images for each propagated hypothesis and obtain the cost matrix
%%%%% formula 8
\begin{equation}
\setlength{\abovedisplayskip}{2pt}
\setlength{\belowdisplayskip}{2pt}
{\textsf M}=
\begin{bmatrix}
{m}_{11} & {m}_{12} & \cdots & {m}_{1N} \\
{m}_{21} & {m}_{22} & \cdots & {m}_{2N} \\
\vdots & \vdots & \ddots & \vdots \\
{m}_{81} & {m}_{82} & \cdots & {m}_{8N}
\end{bmatrix},
\end{equation}
where ${m}_{ij}$ is a matching cost measuring photometric consistency for the ${i}$th hypothesis and ${j}$th view, and $N$ is the number of source images different from the reference. It is worthy of note that, $N$ is preselected according to viewing angle and baseline or sparse 3D points from Structure from Motion (SfM) for large-scale scenes. The construction of cost matrix in checkerboard pattern can not  only be employed to infer good views and hypotheses parallel as shown later, but also not introduce additional calculations.

After the construction of cost matrix, we perform our inference strategy from two dimensions of ``view" and ``hypothesis". First, we observe each column of the cost matrix from the point of ``view". A key observation is that the 8 matching costs are always high for a bad view. In contrast, there are always some corresponding smaller matching costs within the 8 propagated hypotheses for a good view. Furthermore, the matching costs for the good view will decrease with the iteration of our algorithm. Therefore, we define the matching cost threshold function as
%%%%% formula 9
\begin{equation}
\setlength{\abovedisplayskip}{2pt}
\setlength{\belowdisplayskip}{2pt}
{\tau}_{mc}(t)={\tau}_{mc\_init}\cdot e^{-\frac{t^2}{\alpha}},
\end{equation}
where ${t}$ means the ${t}$th iteration and ${\tau}_{mc\_init}$ is the initial matching cost threshold. For each view, if there exist more than ${n}_{1}$ matching costs which are less than ${\tau}_{mc}(t)$ and less than ${n}_{2}$ matching costs which are more than ${\tau}_{up}$, then the view will be considered as a good view. That is, we get the current view selection set $\textrm{S}_{t}$ in the iteration $t$.

Afterwards, we observe each row of the cost matrix from the point of ``hypothesis". To evaluate the importance of each selected view, we need to compute the confidence of its corresponding 8 matching costs. The confidence of a matching cost is calculated as follows,
%%%%% formula 10
\begin{equation}
\setlength{\abovedisplayskip}{2pt}
\setlength{\belowdisplayskip}{2pt}
%C(p,h(p))=e^{-\frac{(1-m(p,h(p)))^2}{2{\alpha}^2}},
C({m}_{ij})=e^{-\frac{{m}_{ij}^2}{2{\beta}^2}}.
\end{equation}
%where $p$ is the corresponding patch at pixel ${\bf x}$ in the reference image, $h(\cdot)$ means the hypothesis induced corresponding patch in a source image and $m$ is the matching cost between them.
It integrates the photometric consistency and makes good views more discriminative. The importance of each selected view can be defined as
%%%%% formula 11
\begin{equation}
\setlength{\abovedisplayskip}{2pt}
\setlength{\belowdisplayskip}{2pt}
{\psi}({\mathcal{X}}^{j}) = \frac{1}{8}{\sum_{i=1}^{8}{C(m_{ij})}}, {\mathcal{X}}^{j}\in\textrm{S}_{t}.
\end{equation}
We average the confidence over all the hypotheses because the propagated hypotheses tend to be consistent with the iteration of our algorithm. Although some of the hypotheses may be quite wrong, they will play the same role in all the selected views. This leads to almost the same weight change in final aggregation. To suppress the bad views, we only remain at most ${k}$ important views, and the weight of others are set to zero.

At the same time, we record the most important view $v_{t}$ in the iteration $t$. In practice, due to the association of the selected view set of two adjacent iterations, we modify the importance function of each selected view as
%%%%% formula 12
\begin{equation}
\setlength{\abovedisplayskip}{2pt}
\setlength{\belowdisplayskip}{2pt}
{\psi}_{mod}({\mathcal{X}}^{j})=\left\{ \begin{array}{ll}
\varepsilon({v}_{t-1}\in\textrm{S}_{t})\cdot{\psi}({\mathcal{X}}^{j}), & {\mathcal{X}}^{j}\in\textrm{S}_{t}; \\
0.2\cdot{\mathbb{I}({v}_{t-1}\in\textrm{S}_{t-1})}, & {\textrm{else.}}
\end{array} \right.
\end{equation}
%%%%% formula 13
\begin{equation}
\setlength{\abovedisplayskip}{2pt}
\setlength{\belowdisplayskip}{2pt}
\varepsilon({v}_{t-1}\in\textrm{S}_{t})=\left\{ \begin{array}{ll}
2, & {\mathbb{I}({v}_{t-1}\in\textrm{S}_{t})=1}; \\
1, & {\textrm{else.}}
\end{array} \right.
\end{equation}
where ${\mathbb{I}(\cdot)}$ is the indicator function. If a view simultaneously satisfys two conditions, including (\romannumeral1) it is the most import view in the ${t-1}$ iteration and (\romannumeral2) it will be selected in the $t$ iteration, it will have the more importance. And a view selected in ${t-1}$ iteration but not in ${t}$ iteration may also be true. This modification can make our view selection method more robust. Finally, we can get the unified cost representation for each hypothesis as follows,
%%%%% formula 14
\begin{equation}
\setlength{\abovedisplayskip}{2pt}
\setlength{\belowdisplayskip}{2pt}
m_{final}(i)=\frac{\sum{{\psi}_{mod}(\mathcal{X}^{j})}\cdot{m}_{\cdot j}}{\sum{{\psi}_{mod}(\mathcal{X}^{j})}}.
\end{equation}
Thus, the current optimal hypothesis at pixel ${\bf x}$ in the reference image ${{\mathcal{X}}}^{ref}$ is determined by selecting the minimal $m_{final}$ from the 8 propagated hypotheses.

Our proposed multi-hypothesis joint strategy makes the view selection has following characteristics: (\romannumeral1) all the hypotheses in the same iteration have the same matching cost comparison baseline, that is, it have the same selected views for 8 hypotheses, (\romannumeral2) the selected views are discriminative via our designed confidence integration, (\romannumeral3) the number of selected views is adaptively determined. Thanks to the multi-hypothesis joint view selection, the resulting 3D point clouds are more accurate and complete (as shown in section~\ref{ED}). Moreover, it breaks through the limitation of strong pixel dependence in a sequential inference mode, and can still remain the massively parallel computing power.

\subsection{Implementation Details}
In this section, we detail the matching cost computation and refinement of our multi-view stereo method. {\em Gipuma} employs the same matching cost as Bleyer {\em et al.}~\cite{Alpher13} do. Although it is a weighted way, it can still not handle the boundary details well because it dose not contain the spatial distance information~\cite{Alpher30}. In our multi-view stereo algorithm, we adopt a bilaterally weighted adaption of NCC to compute the matching cost as Sch{\"o}nberger {\em et al.}~\cite{Alpher10} do, which is defined as
%%%%% formula 15
\begin{equation}
\setlength{\abovedisplayskip}{2pt}
\setlength{\belowdisplayskip}{2pt}
m(p,h(p))=1-\frac{cov(p,h(p))}{\sqrt{cov(p,p)cov(h(p),h(p))}},
\end{equation}
where $p$ is the corresponding patch at pixel ${\bf x}$ in the reference image, $h(\cdot)$ means the patch induced by the corresponding hypothesis in a source image, $cov(x,y)=E(x-E(x))E(y-E(y))$ is the weighted covariance and $E(x)=\sum_{i}{{w_i}{x_i}}/\sum_{i}{w_i}$ is the weighted average. The per-pixel weight equals to $w_i=exp(-\frac{\nabla{I}}{2{\sigma}_{I}^2}-\frac{\nabla{x}}{2{\sigma}_{x}^2})$, where ${\nabla I}$ means the color difference and ${\nabla x}$ means the spatial distance. Meanwhile, to speed up the matching cost computation, we adopt a skipping way of using only every other row and column in the window to execute the computation~\cite{Alpher09}.

The refinement in the {\em PatchMatch} based stereo plays a role of increasing the diversity of hypotheses and fine-tuning the depth and normal. As analysed in ~\cite{Alpher10}, the current best depth and normal parameters have the following states: neither of them, one of them, or both of them have the optimal solutions or are close to it. Hence, we exploit their refinement method, referring to ~\cite{Alpher10} for more details. Unlike their integration, we enforce bisection refinement in the first 3 iterations, and combine both the bisection and the above refinement method in the rest of the iterations to make a tradeoff between accuracy and efficiency. Finally, we apply a median filter of size $4\times4$ as a postprocessing in our depth map generation.

To reduce the influence of wrong depth estimation and noise, we enforce a straightforward consistency check to fuse the depth maps based on the estimated depth and normal as done in ~\cite{Alpher09}. The main difference is Galliani {\em et al.} struggle to simultaneously maintain the accuracy and completeness of 3D models due to their slightly deficient depth map estimation. This makes them need to set different fusion parameters according to different applications. In all our experiments, these parameters are fixed with our more accurate and complete depth map estimation (See Section~\ref{ED}). Therefore, the threshold for disparity difference $f_{\epsilon}$, normal difference $f_{ang}$ and number of consistent source images $f_{con}$ in ~\cite{Alpher09} is set to 0.3, 30 and 2 respectively.

\section{Experiments and Discussions}\label{ED}
In this section, we evaluate our proposed AMHMVS algorithm on different representative public benchmarks, including Strecha datasets~\cite{Alpher15} and ETH3D benchmark~\cite{Alpher31}.
Besides, we also show some dense reconstruction results of unstructured Internet photo collections.
All of our experiments are conducted by using C++ with a CUDA implementation on a machine with two Intel Xeon CPU E5-2630 v3 2.40GHz, and two Nvidia GeForce GTX TitanX GPUs with 12GB global memory.

Throughout the experiments, we use the following parameters. The number of iterations is set to 6. For the matching cost threshold function, ${\tau}_{mc\_init}$ is set to 0.8 and ${\alpha}$ is set to 90.0. We make ${\beta}$ equal to 0.3 to compute the hypothesis' confidence. $n_{1}$, $n_{2}$ and ${\tau}_{up}$ is set to 2, 3 and 1.2 respectively. We set $k$ to 4 to select the good views. And we choose ${\sigma}_{I}=3.0$ and ${\sigma}_{x}=30.0$ in the computation of matching cost.

\subsection{Depth Map Evaluation: Strecha Benchmark}
Since our method is depth map based multi-view stereo, we will evaluate the accuracy and completeness of generated depth maps in this section. Strecha Benchmark~\cite{Alpher15} comprises 6 outdoor datasets with ${3072\times2048}$ resolution, as well as ground truth 3D models captured by a laser scanner. However, it only includes two datasets, Fountain-P11 and Herjzesu-P9, which provide the ground truth depth map measurements. Following ~\cite{Alpher19}, we calculate the number of pixels with a depth error less than 2${cm}$ and 10${cm}$ from the ground truth and also omit the evaluation of the dataset's two extremal views as done in ~\cite{Alpher19}.

%figure 4
\begin{figure}[t]
\begin{center}
   %\includegraphics[width=0.315\linewidth]{fig4//Fountain//SCP//depth2000.png}
%   \includegraphics[width=0.315\linewidth]{fig4//Fountain//MHJ//depth2000.png}
%   \includegraphics[width=0.315\linewidth]{fig4//Fountain//AMH//depth2000.png}
%   \includegraphics[width=0.315\linewidth]{fig4//Herzjesu//SCP//depth2000.png}
%   \includegraphics[width=0.315\linewidth]{fig4//Herzjesu//MHJ//depth2000.png}
%   \includegraphics[width=0.315\linewidth]{fig4//Herzjesu//AMH//depth2000.png}
   \includegraphics[width=0.24\linewidth]{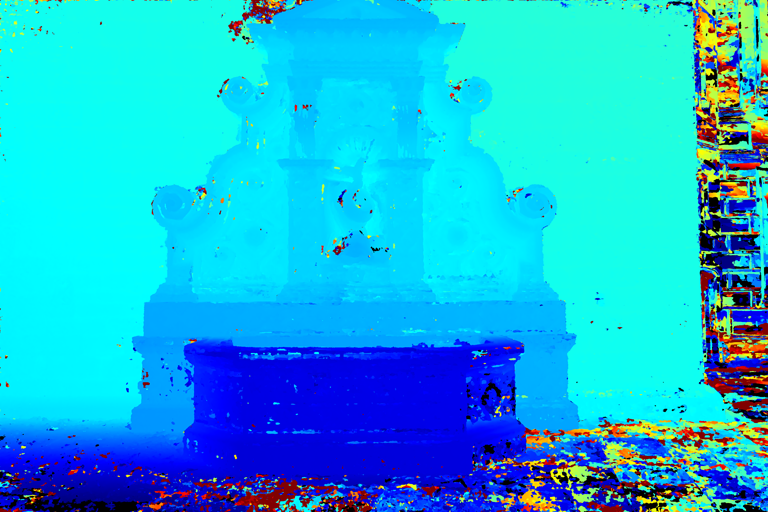}
   \includegraphics[width=0.24\linewidth]{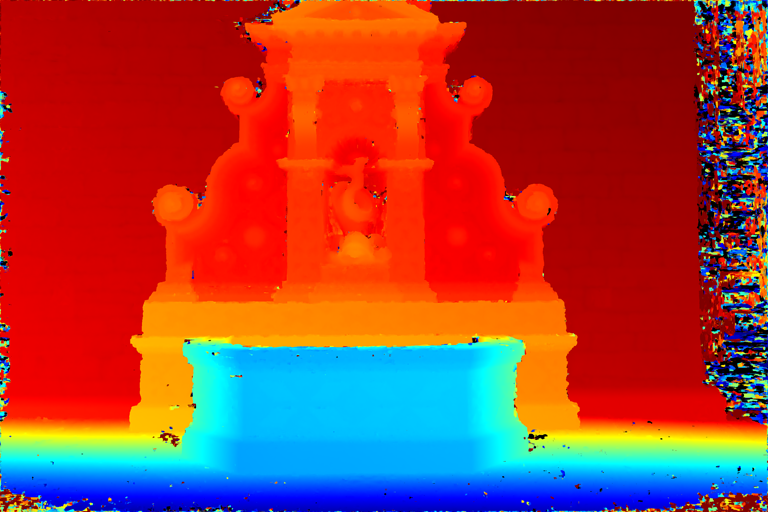}
   \includegraphics[width=0.24\linewidth]{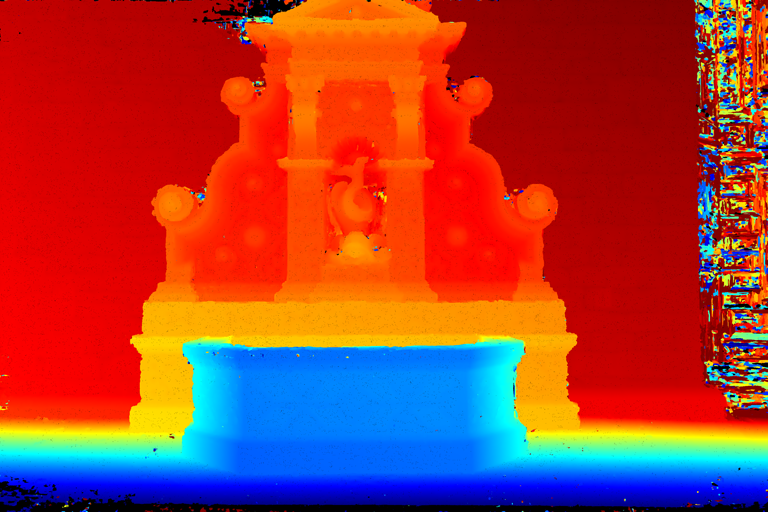}
   \includegraphics[width=0.24\linewidth]{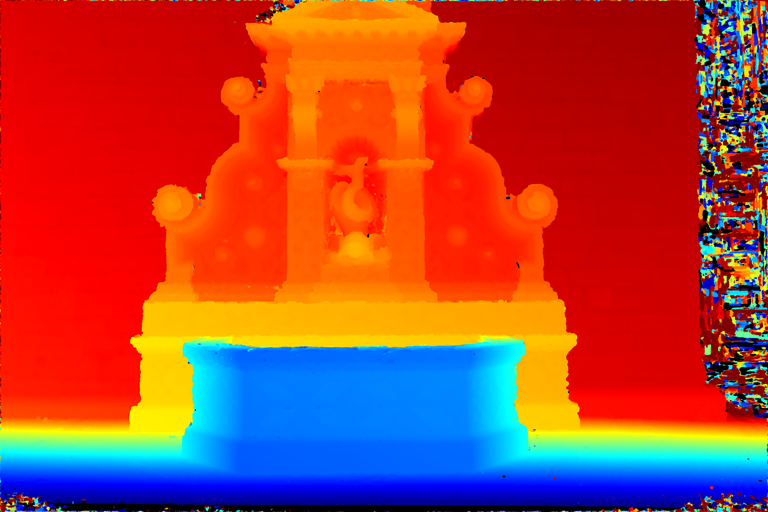}
   \includegraphics[width=0.24\linewidth]{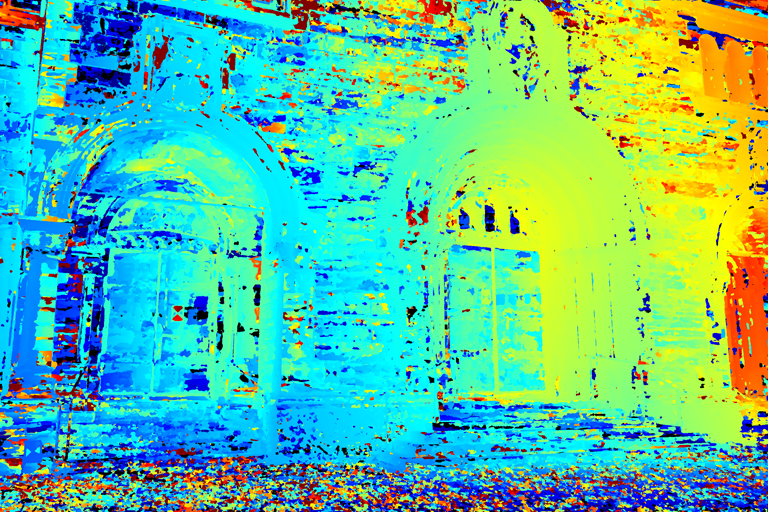}
   \includegraphics[width=0.24\linewidth]{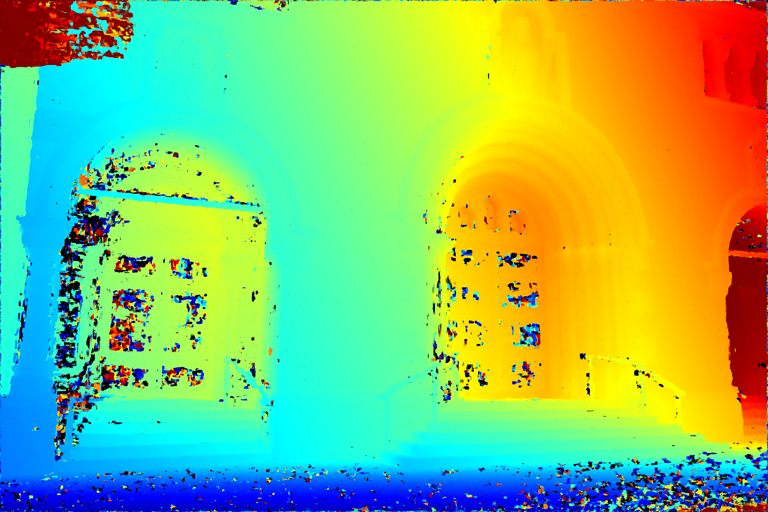}
   \includegraphics[width=0.24\linewidth]{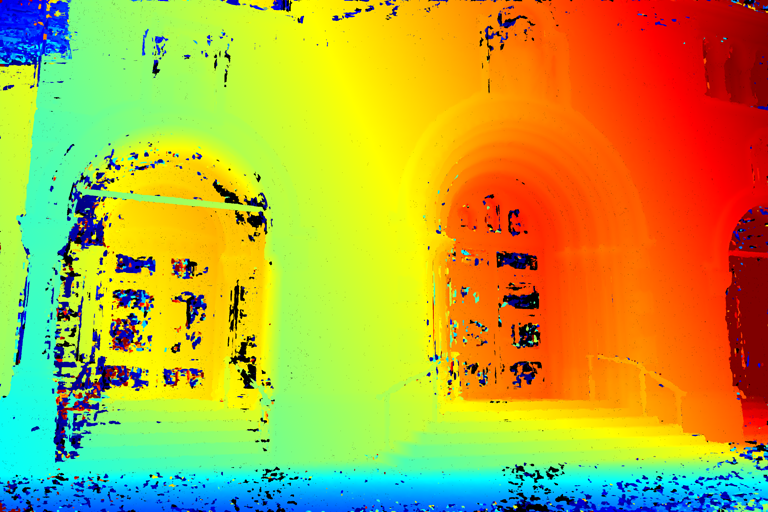}
   \includegraphics[width=0.24\linewidth]{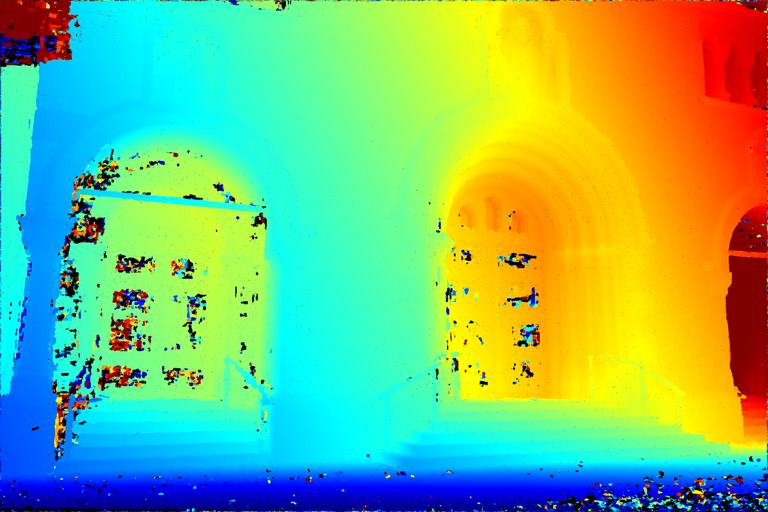}
   \qquad\qquad\qquad\qquad\textrm{(a)}\qquad\qquad\qquad\qquad\quad\textrm{(b)}\qquad\qquad\qquad\quad\quad\textrm{(c)}\qquad\qquad\qquad\quad\quad\textrm{(d)}\qquad\qquad
\end{center}
   \caption{The generated depth maps for Strecha Benchmark~\cite{Alpher15}. The results of the first row are the depth maps of Fountain. The results of the second row are the depth maps of Herzjesu. (a) Generated depth maps by Gipuma~\cite{Alpher09}. (b) Generated depth maps w/o Asymmetric Checkerboard Propagation. (c) Generated depth maps w/o Multi-Hypothesis Joint View Selection. (d) Generated depth maps with both the above proposed strategies (Best viewed in color). }
\label{fig:depthmap}
\end{figure}

To evaluate the benefit from our proposed asymmetric checkerboard propagation and multi-hypothesis joint view selection, we carry out the following ablation experiments: (\romannumeral1) the original Gipuma algorithm, (\romannumeral2) our method using symmetric checkerboard propagation instead of asymmetric checkerboard propagation, (\romannumeral3) our method using heuristic view selection instead of multi-hypothesis joint view selection, and (\romannumeral4) our method combing both our proposed strategies. We show the qualitative results in Fig.~\ref{fig:depthmap}. We see that depth maps generated by the original Gipuma are very coarse and not consistent. By using our method combined with only asymmetric checkerboard propagation or multi-hypothesis view selection, the generated depth maps are greatly improved. However, when using only one of our proposed strategies, the depth estimation in ground regions are not very smooth, and the boundaries are with much noise. It is because that, with symmetric checkerboard propagation, many regions can not obtain good hypotheses' propagation until the iteration terminates. Our method exploits the confidence-guided asymmetric checkerboard propagation to make good hypotheses expand further. Because the true hypotheses are always submerged by the false hypotheses in heuristic view selection, the generated depth maps are with many holes. Our proposed multi-hypothesis view selection can better infer the aggregated view subset for every pixel to eliminate the interference of false hypotheses.

\begin{figure}[t]
\begin{center}
   %\includegraphics[width=0.8\linewidth]{egfigure.eps}
   \includegraphics[width=0.48\linewidth,height=0.40\linewidth]{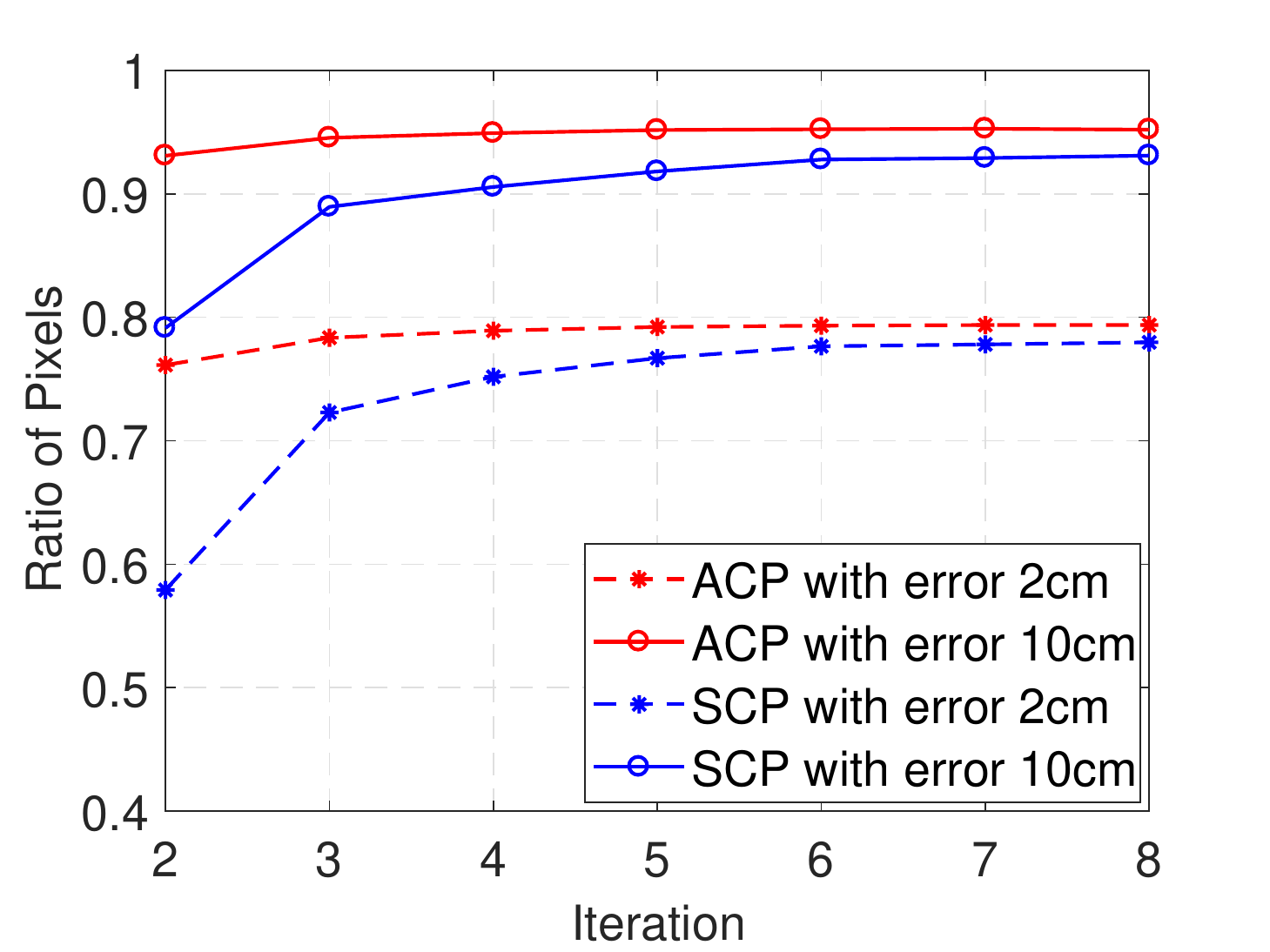}
   \includegraphics[width=0.48\linewidth,height=0.40\linewidth]{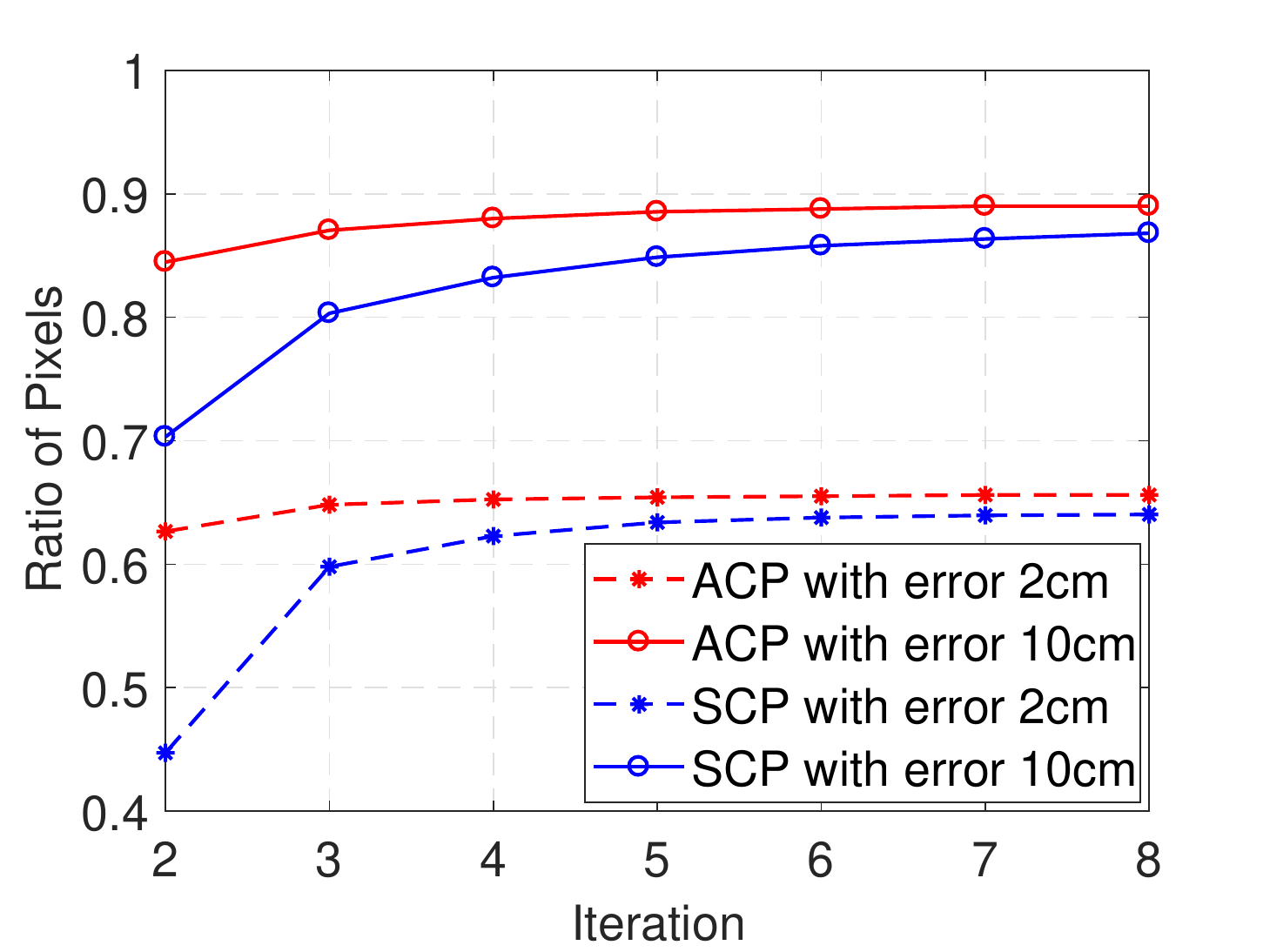}
   \qquad\textrm{(a)}\qquad\qquad\qquad\qquad\qquad\qquad\qquad\qquad\qquad\textrm{(b)}\qquad
\end{center}
   \caption{The performance comparison between Asymmetric Checkerboard Propagation (ACP) and Symmetric Checkerboard Propagation (SCP): (a) Accuracy and completeness for increasing number of iterations on Fountain dataset (b) Accuracy and completeness for increasing number of iterations on Herzjesu dataset.}
\label{fig:ACPVSSCP}
\label{fig:onecol}
\end{figure}

Next, We verify the convergence and efficiency of our propagation strategy. As shown in Fig.~\ref{fig:ACPVSSCP}, with the asymmetric checkerboard propagation scheme,
%the ratio of pixels with absolute error less than 2${cm}$ and 10${cm}$ is much higher than the symmetric checkerboard propagation.
the ratio of accurate pixels is much higher than the symmetric checkerboard propagation.
It is because that the asymmetric checkerboard propagation breaks through the restriction of the original limited propagation distance. Not only can it make the good hypotheses expand further, but it can also generate more hypotheses in a sense. Moreover, our proposed propagation scheme can achieve a stable state more quickly, which can further reduce the runtime of depth map generation, that is, we can obtain the satisfactory results with fewer iterations. Therefore, we set the number of iterations to 6 in all our experiments.

%figure 3

\begin{table*}[t]
  \caption{Percentage of pixels with absolute errors below 2cm and 10cm on Strecha Benchmark~\cite{Alpher15}. The related values are from ~\cite{Alpher19,Alpher08,Alpher10}. Ours (AMHMVS) w/o asymmetric checkerboard propagation ($\backslash$ACP) instead of symmetric diffusion-like propagation, w/o multi-hypothesis joint view selection ($\backslash$MHJ) instead of heuristical view selection method and with both the proposed strategies.}
  \begin{center}
  \begin{tabular}{|c|c|c|c|c|c|c|c|c|c|c|c|c|c|c|}
  \hline
  & error & \cite{Alpher08} & \cite{Alpher19} & \cite{Alpher01} & \cite{Alpher32} & \cite{Alpher33} & \cite{Alpher20} & \cite{Alpher09} & \cite{Alpher10} & $\backslash$ACP & $\backslash$MHJ & {\bf Ours}  \\
  \hline
  \multirow{2}{*}{Fountain} & $2 cm$ & 0.769 & 0.754 & 0.731 & 0.712 & 0.732 & 0.824 & 0.693 & 0.827 & 0.777 & 0.783 & 0.793 \\
  \cline{2-13}
  & $10 cm$ & 0.929 & 0.930 & 0.838 & 0.832 & 0.822 & 0.973 & 0.838 & 0.975 & 0.928 & 0.938 & 0.952 \\
  \hline
  \multirow{2}{*}{Herzjesu} & $2 cm$ & 0.650 & 0.649 & 0.646 & 0.220 & 0.658 & 0.739 & 0.283 & 0.691 & 0.638 & 0.628 & 0.655 \\
  \cline{2-13}
  & $10 cm$ & 0.844 & 0.848 & 0.836 & 0.501 & 0.852 & 0.923 & 0.455 & 0.931 & 0.858 & 0.853 & 0.888 \\
  \hline
  \end{tabular}
  \end{center}
  \label{tab:depthmap}
\end{table*}

Furthermore, we list the detail comparison
%percentage of pixels with absolute error below 2${cm}$ and 10${cm}$
of other methods and ours in Table~\ref{tab:depthmap}. We show that our method is competitive with other methods in accuracy. Through the above comparison of depth maps between Gipuma and ours, it is not surprising that our results are much better than Gipuma. Except for CMPMVS~\cite{Alpher20} and COLMAP~\cite{Alpher10}, our method is better than others. The results of CMPMVS are obtained by constructing the s-t graph to fuse multiple depth maps, while our method individually estimates every depth map. As for the results generated by COLMAP, it first employs the photometric information to estimate a intermediate depth map for each image, and then leverages the global information from all intermediate depth maps to check the geometric consistency. It means that COLMAP integrates the consistency check in the depth map generation stage. In contrast, our results are got without consistency check while are a little bit inferior to COLMAP. The consistency check of our method is embedded in the fusion stage. In next section, we will further present the evaluation of our reconstructed point cloud by combing the consistency check.

\begin{figure*}[t] %[b]
\begin{center}
%\fbox{\rule{0pt}{2in} \rule{.9\linewidth}{0pt}}
   \includegraphics[width=0.98\linewidth]{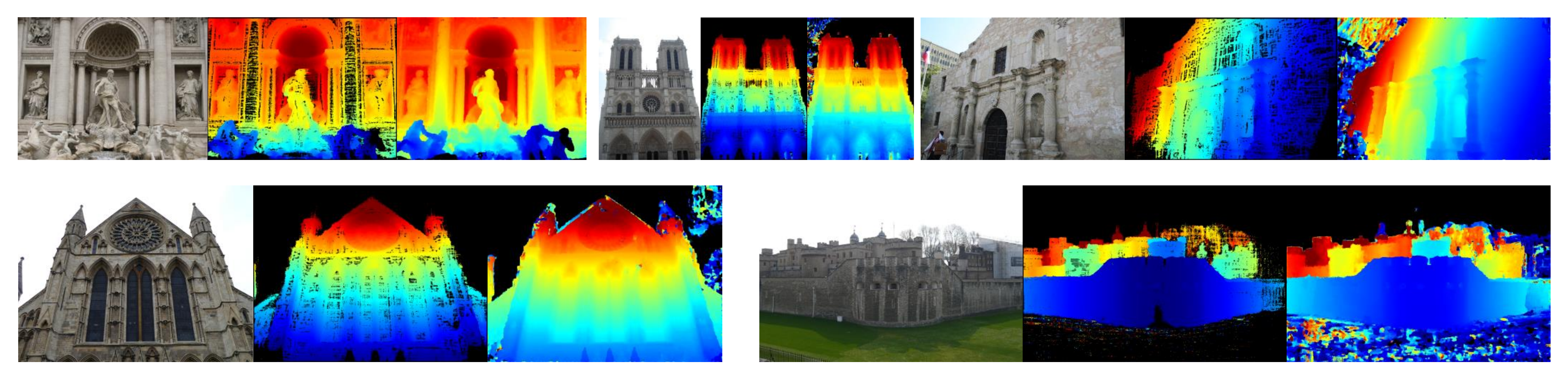}
\end{center}
   \caption{Each image triplet depicts a reference image along with COLMAP's and our depthmap estimation (Best viewed in color). Note that, the depth maps obtained by COLMAP have been filtered so that its depth ranges are small than ours. It causes the color representation of our depth map is a bit different from COLMAP's.}
\label{fig:IPC}
\end{figure*}

We also test our method on Internet photo collections from Kyle Wilson and Noah Snavely~\cite{Alpher36} for six different scenes: Fontana di Trevi, Notre Dame de Paris, Alamo, Yorkminster and Tower of London. We resize the imagery to no more than 1024 pixels for each dimension. Camera poses are calculated by using Bundler~\cite{Alpher23} with PBA~\cite{Alpher37}. We run COLMAP on these datasets with its default settings, except 3 iterations for photometric consistency and 2 iterations for geometric consistency. Moreover, we adopt the skipping way to compute the bilateral weighted NCC for COLMAP. The estimated depth maps are shown in Fig.~\ref{fig:IPC}. The results illustrate that, both method can get consistent depth map estimation on these challenging unstructured Internet photo collections. Due to the early strong filtering in depth map estimation, the depth maps obtained by COLMAP have more holes than ours. It may result in decreased completeness in depth map fusion, which can be seen in Section~\ref{PCE}.

\subsection{Point Cloud Evaluation: ETH3D Benchmark}\label{PCE}

In this section, we evaluate our method on 12 high-resolution (${6048\times4032}$) test datasets of ETH3D Benchmark~\cite{Alpher31}. ETH3D Benchmark covers a diverse set of viewpoints and scene types, ranging from natural scenes to man-made indoor and outdoor environments.
%with unprecedented high-resolution (${6048\times4032}$) in MVS.
%The scene structure is relatively complex, including the intractable weakly textured surfaces and thin structures, which provides a challenge to our method.
In our point cloud evaluation, we employ a pair of GPUs and resize the image to 3200 to test the efficiency of our method,
 which is the same as the settings of COLMAP~\cite{Alpher10}. To meet the actual needs, we also leverage the sparse 3D points to reduce the view selection scale. It means that we set $N$ to 20 in our point cloud evaluation.
% which means the source images are considered that have shared sparse 3D points with the reference image.
 In point cloud evaluation, we evaluate our reconstructed points in terms of accuracy, completeness and composite index $F_{1}$ score.

 \begin{figure*}[t]
\begin{center}
   %\includegraphics[width=0.24\linewidth]{fig5//court//snapshot00.png}
%   \includegraphics[width=0.24\linewidth]{fig5//deliv//snapshot00.png}
%   \includegraphics[width=0.24\linewidth]{fig5//relief//snapshot01.png}
%   \includegraphics[width=0.24\linewidth]{fig5//terra//snapshot01.png}
   %\includegraphics[width=0.24\linewidth]{fig5//court//snapshot01.png}
%   \includegraphics[width=0.24\linewidth]{fig5//deliv//snapshot01.png}
%   \includegraphics[width=0.24\linewidth]{fig5//relief//snapshot00.png}
%   \includegraphics[width=0.24\linewidth]{fig5//terra//snapshot02.png}
   \includegraphics[width=0.192\linewidth]{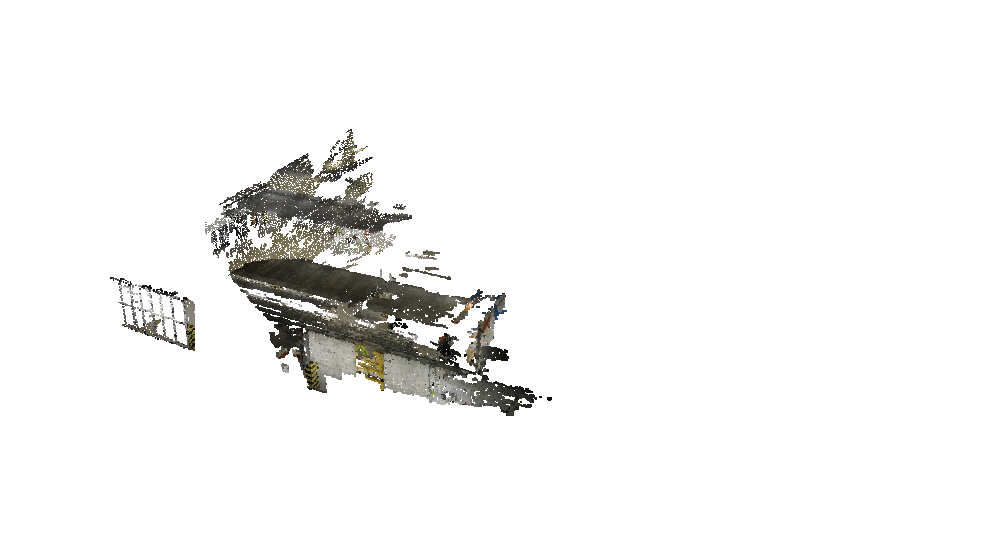}
   \includegraphics[width=0.192\linewidth]{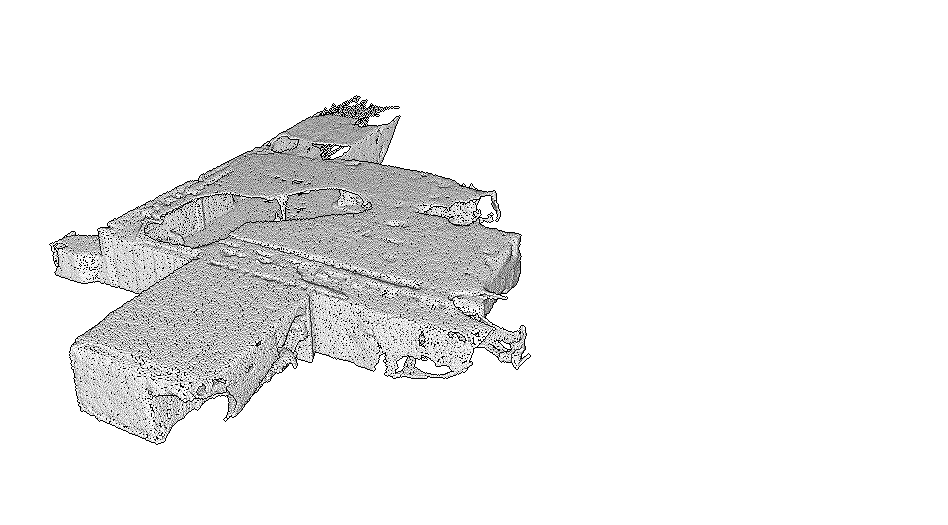}
   \includegraphics[width=0.192\linewidth]{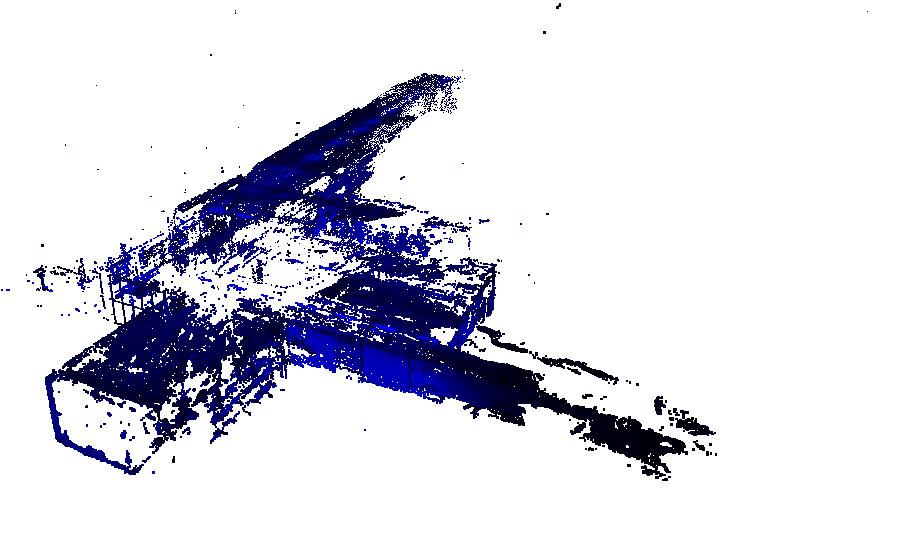}
   \includegraphics[width=0.192\linewidth]{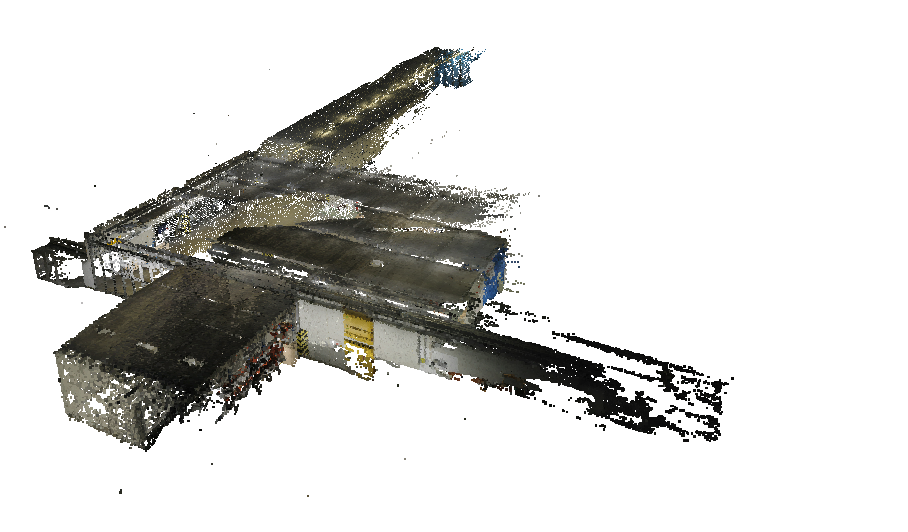}
   \includegraphics[width=0.192\linewidth]{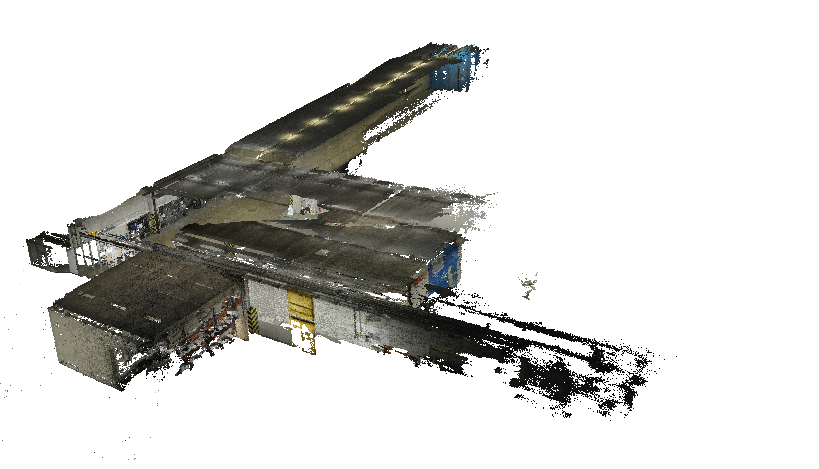}
   \qquad \textrm{(a) PMVS} \qquad\quad \textrm{(b) CMPMVS} \qquad \textrm{(c) Gipuma} \qquad \textrm{(d) COLMAP} \qquad \textrm{(d) AMHMVS} \quad
\end{center}
   \caption{The dense 3D reconstructions of delivery area dataset for other methods, including PMVS~\cite{Alpher01}, CMPMVS~\cite{Alpher20}, Gipuma~\cite{Alpher09}, COLMAP~\cite{Alpher10} and ours (AMHMVS). The related results are from ~\cite{Alpher34}. }
\label{fig:seth3d}
\end{figure*}

 The aim of multi-view stereo is to obtain the dense 3D models. Consistency check, which checks the depth coherence among multiple images, is utilized in our fusion stage to improve the accuracy of depth estimation. As we can see in Fig.~\ref{fig:seth3d}, by combining the consistency check with our well estimated depth maps, our method gets better results  than others in 3D reconstructions, especially in completeness.
%Our algorithm can maintain the detail structure of different scenes including the outdoor and indoor.

\begin{table}
  %\footnotesize
  \scriptsize
  %\tiny
  \caption{ETH3D Benchmark evaluation in 12 high-resolution test multi-view scenario showing accuracy / completeness / $F_{1}$ score (in \%) at different thresholds ($t$) for other methods, including PMVS~\cite{Alpher01}, CMPMVS~\cite{Alpher20}, Gipuma~\cite{Alpher09}, COLMAP~\cite{Alpher10} and ours (AMHMVS). The last row shows the runtime of these methods. The related values are from ~\cite{Alpher34} (Ind. means indoor while Outd. means outdoor).}
  \begin{center}
  \begin{tabular}{c|c|c|c|c|c|c}
  \hline
  Type & $t$[cm] & PMVS~\cite{Alpher01} & CMPMVS~\cite{Alpher20} & Gipuma~\cite{Alpher09} & COLMAP~\cite{Alpher10} & {\bf Ours (AMHMVS)}  \\
  \hline
  \multirow{6}{*}{Ind.} & $1$ & 83.15 / 22.43 / 33.29 & 60.89 / 52.64 / 55.97 & 72.83 / 24.26 / 31.91 & $\textbf{84.11}$ / 48.05 / 58.81 & 80.11 / $\textbf{55.94}$ / $\textbf{65.20}$ \\
  & $2$ & 90.66 / 28.16 / 40.28 & 73.57 / 64.41 / 68.16 & 86.33 / 31.44 / 41.86 & $\textbf{91.97}$ / 59.65 / 70.41 & 91.14 / $\textbf{64.81}$ / $\textbf{73.93}$ \\
  & $5$ & 95.33 / 35.53 / 48.46 & 85.67 / 74.56 / 79.20 & 95.80 / 42.05 / 54.91 & 96.62 / 73.00 / 89.74 & $\textbf{97.43}$ / $\textbf{74.96}$ / $\textbf{83.68}$ \\
  & $10$ & 96.97 / 42.50 / 55.40 & 91.39 / 80.11 / 84.92 & 98.31 / 52.22 / 65.41 & 98.11 / $\textbf{82.82}$ / 89.28 & $\textbf{98.76}$ / 82.61 / $\textbf{89.42}$ \\
  & $20$ & 97.94 / 51.34 / 63.57 & 95.44 / 85.14 / 89.67 & 99.16 / 65.00 / 76.75 & $\textbf{99.25}$ / $\textbf{98.85}$ / $\textbf{94.87}$ & 99.24 / 89.63 / 93.95 \\
  & $50$ & 98.81 / 65.05 / 74.97 & 98.19 / 92.04 / 94.90 & $\textbf{99.59}$ / 83.50 / 90.15 & 99.37 / $\textbf{98.17}$ / $\textbf{98.75}$ & 99.54 / 96.62 / 97.99 \\
  \hline
  \multirow{6}{*}{Outd.} & $1$ & 77.36 / 32.99 / 45.02 & 63.52 / 63.46 / 57.81 & 59.70 / 37.12 / 43.33 & $\textbf{82.68}$ / 59.47 / 68.64 & 69.68 / $\textbf{71.28}$ / $\textbf{70.07}$ \\
  & $2$ & 88.34 / 42.89 / 55.82 & 79.77 / 73.84 / 76.28 & 78.78 / 45.30 / 55.16 & $\textbf{92.04}$ / 72.98 / 80.81 & 83.96 / $\textbf{80.03}$ / $\textbf{81.77}$ \\
  & $5$ & 93.91 / 50.18 / 63.48 & 92.09 / 79.17 / 84.48 & 93.81 / 54.30 / 67.24 & $\textbf{97.13}$ / 83.94 / 89.74 & 94.07 / $\textbf{87.10}$ / $\textbf{90.39}$ \\
  & $10$ & 95.95 / 55.17 / 68.12 & 96.77 / 81.59 / 87.74 & 97.36 / 62.40 / 75.18 & $\textbf{98.64}$ / 89.70 / 93.79 & 97.51 / $\textbf{90.57}$ / $\textbf{93.87}$ \\
  & $20$ & 97.65 / 60.97 / 73.09 & 98.84 / 83.57 / 89.78 & 98.64 / 72.79 / 83.38 & $\textbf{99.41}$ / $\textbf{94.25}$ / $\textbf{96.71}$ & 99.26 / 85.02 / 96.34 \\
  & $50$ & 98.85 / 68.75 / 79.01 & 99.41 / 92.04 / 94.13 & 99.49 / 88.40 / 93.51 & 99.70 / $\textbf{98.65}$ / $\textbf{99.17}$ & $\textbf{99.72}$ / 97.21 / 98.44 \\
  \hline
  \multirow{6}{*}{All} & $1$ & 81.70 / 25.07 / 36.22 & 61.54 / 55.34 / 57.81 & 69.55 / 27.47 / 34.77 & $\textbf{83.75}$ / 50.90 / 61.27 & 77.50 / $\textbf{59.78}$ / $\textbf{65.20}$ \\
  & $2$ & 90.08 / 31.84 / 44.16 & 75.12 / 66.77 / 70.19 & 84.44 / 34.91 / 45.18 & $\textbf{91.97}$ / 62.98 / 73.01 & 89.34 / $\textbf{68.62}$ / $\textbf{75.89}$ \\
  & $5$ & 94.97 / 39.19 / 52.22 & 87.28 / 75.71 / 80.52 & 95.31 / 45.11 / 57.99 & $\textbf{96.75}$ / 75.74 / 83.96 & 96.59 / $\textbf{77.99}$ / $\textbf{85.36}$ \\
  & $10$ & 96.71 / 45.67 / 58.58 & 92.74 / 80.48 / 85.62 & 98.07 / 54.77 / 67.86 & 98.25 / 84.54/ 90.40 & $\textbf{98.44}$ / $\textbf{84.60}$  / $\textbf{90.53}$ \\
  & $20$ & 97.86 / 53.75 / 65.95 & 96.29 / 84.75 / 89.70 & 99.03 / 66.95 / 78.40 & 98.99 / $\textbf{92.18}$ / $\textbf{95.33}$ & $\textbf{99.25}$ / 90.64 / 94.55 \\
  & $50$ & 98.82 / 65.98 / 75.98 & 98.50 / 90.65 / 94.13 & 99.57 / 84.73 / 90.99 & 99.45 / $\textbf{98.29}$ / $\textbf{98.86}$ & $\textbf{99.59}$ / 96.77 / 98.10 \\
  \hline
  Time & -- & 957.08s & 1983.08s & 689.75s & 1658.33s & 967.92s \\
  \hline
  \end{tabular}
  \end{center}
  \label{tab:pc_ioa}
\end{table}

\begin{figure*}[t]
\begin{center}
%\fbox{\rule{0pt}{2in} \rule{.9\linewidth}{0pt}}
   \includegraphics[width=0.98\linewidth]{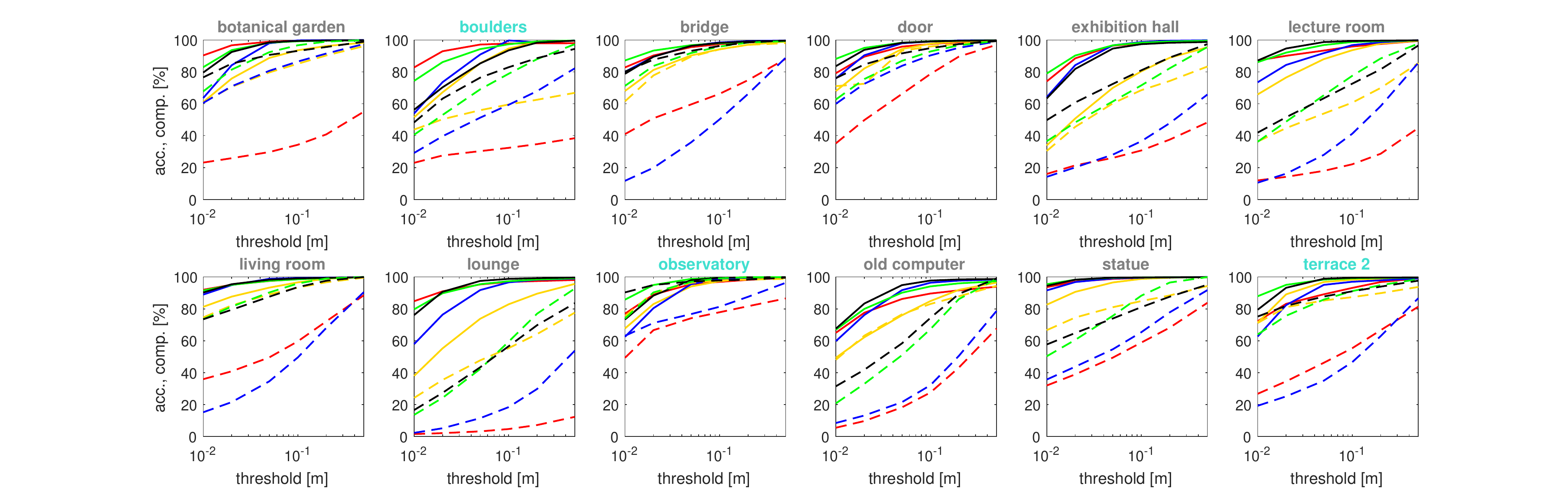}
\end{center}
   \caption{ETH3D Benchmark evaluation in the high-resolution multi-view scenario (\textcolor[rgb]{0.5,0.5,0.5}{indoor} and \textcolor[rgb]{0.25,0.875,0.8125}{outdoor} datasets) for different methods, including \textcolor[rgb]{1,0,0}{PMVS}~\cite{Alpher01}, \textcolor[rgb]{1,0.84,0}{CMPMVS}~\cite{Alpher20}, \textcolor[rgb]{0,0,1}{Gipuma}~\cite{Alpher09}, \textcolor[rgb]{0,1,0}{COLMAP}~\cite{Alpher10} and \textcolor[rgb]{0,0,0}{ours (AMHMVS)}. Results are shown as a solid line for accuracy and as a dashed line for completeness. The related values are from ~\cite{Alpher34} (Best viewed in color).}
\label{fig:eth3db}
\end{figure*}

The specific performance of ours and other methods, including PMVS~\cite{Alpher01}, CMPMVS~\cite{Alpher20}, Gipuma~\cite{Alpher09} and COLMAP~\cite{Alpher10}, are summarized in Table~\ref{tab:pc_ioa}, and the details for each dataset are shown in Fig.~\ref{fig:eth3db}\footnote{More visualized 3D models and details can be seen from \url{https://www.eth3d.net/}.}. In the situation of high precision ({\em i.e.}, the evaluation in the threshold of 1cm, 2cm and 5cm), our method achieves the highest completeness and $F_{1}$ score  while COLMAP achieves the highest accuracy. We notice that COLMAP gets highest completeness and $F_{1}$ score in large distance threshold and ours gets almost highest accuracy in this situation. Because COLMAP has exploited the filtering and geometric consistency check in the depth map generation stage, the effect of its fusion is conservative. However, our results are further boosted by combining consistency check in the final 3D dense reconstruction. Due to the filtering in depth map generation and the further fusion, COLMAP in fact conducts two phase filtering, which causes its final relatively sparse 3D models. This explains why our method is more complete than COLMAP in high precision situation.

We next analysis the performance of CMPMVS, PMVS and Gipuma. Though CMPMVS gets better results in the evaluation of depth maps, its reconstruction results are worse than ours. It is because that CMPMVS has used the global information from multiple depth map before and can not further improve the reconstructed models here. The accuracy of results of PMVS and Gipuma is competitive with ours and COLMAP while the completeness is lower than ours and COLMAP. It is because that PMVS expands the patches from the reliable seeds and stops in the weakly textured surfaces, which results in many regions without depth information. However, the depth in these regions can be estimated in fact. Due to the inaccurate depth estimation for each image by Gipuma, it need a strong consistency check to ensure the accuracy of final models, which means that it only keep the accurate but very few 3D points.

\subsection{Efficiency Analysis}
We list the runtime of other methods and ours in Table~\ref{tab:pc_ioa}. These methods can be divided into two categories: CPU-based and GPU-based. We first discuss the efficiency of CPU-based method, PMVS, and ours. PMVS expands the patches from the reliable seeds and stops early in the weakly textured regions, which reduces a lot of computation. However, our method searches the correspondence for every pixel and its runtime is almost the same as PMVS but gets much better reconstruction results.

 We test the GPU based methods on Strecha datasets in our same platform for the fairness of comparison. Since COLMAP is the representative of PatchMatch based multi-view stereo methods, we run COLMAP~\cite{Alpher10} with the settings stated in Internet photo collections test. To further demonstrate the massively parallel capability of our method, we downsample the images with a factor $\eta$ (we set $\eta$ to 0.5 and 0.25 in our experiments). Due to the depth fusion strategy is run on CPU, we only list the runtime of depth map generation in Table~\ref{tab:depthtime} by a single GPU.

%Our Method Is 6+ Times Faster Than Colmap In The Depth Map Generation Stage, And Around 5 Times Faster Than Colmap In Total. Furthermore, The Efficiency Of Our Method Is Competitive With Gipuma, Which Is The Fastest Method In Mvs. We Should Note That When Our Method Is Compared With Gipuma On The Two Datasets, The Runtime Results Are A Little Inconsistent. It Is Because That Gipuma Always Chooses The Same Number Of Views To Aggregate The Matching Cost, While Our Method Adaptively Selects The True Aggregated Views, Which Makes Our Method Choose Less Aggregated Views Than Gipuma In Herzjesu Dataset For Its Actual Big Change Of Perspective. And Our Method Spends More Time In Fusion Stage Than Gipuma Because We Get More Accurate Pixels To Tackle With.

We see that our method gets different speedup performance with different downsampling factors. Especially, our method can nearly produce a depth map per second when images are downsampled with max image dimension around 800 while COLMAP costs around 8 seconds. This is because our method can utilize all the threads of GPU in different downsampling factors by the checkerboard propagation while the parallel computing capability of COLMAP is limited by the image dimension with its sequential inference model. Moreover, the runtime of COLMAP is proportional to the downsampling factor. This demonstrate the sequential nature of COLMAP's propagation. In fact, there are a lot of images with different dimensions from different benchmarks, images captured by users themselves and especially Internet photo collections. Thus, a multi-view stereo method with a better compatibility of massively parallel power is needed. Our method is truly massively parallel, and it is potential in real environments for its competitive accuracy and completeness.

%\begin{table}
%  \small
%  \caption{Runtime (in second) of other {\em PatchMatch} based methods and ours (AMHMVS) on Strecha Benchmark~\cite{Alpher15}. The results are counted from three aspects: depth map generation stage, fusion stage and total.}
%  \begin{center}
%  \begin{tabular}{|c|c|c|c|c|}
%  \hline
%  & Method & Depth & Fusion & Total \\
%  \hline
%  \multirow{3}{*}{Fountain} & COLMAP~\cite{Alpher10} & 2218 & 44 & 2265 \\
%  \cline{2-5}
%  & Gipuma~\cite{Alpher09} & 250 & 94 & 349 \\
%  \cline{2-5}
%  & {\bf Ours} & 344 & 129 & 473 \\
%  \hline
%  \multirow{3}{*}{Herzjesu} & COLMAP~\cite{Alpher10} & 1498 & 26 & 1524 \\
%  \cline{2-5}
%  & Gipuma~\cite{Alpher09} & 240 & 23 & 263 \\
%  \cline{2-5}
%  & {\bf Ours} & 196 & 76 & 272 \\
%  \hline
%  \end{tabular}
%  \end{center}
%  \label{tab:depthtime}
%\end{table}

\begin{table}
  \small
  \caption{Runtime (in second) of depth map generation for COLMAP~\cite{Alpher10} and ours (AMHMVS) on Strecha Benchmark~\cite{Alpher15} (The original image size is 3072 $\times$ 2048). Note that, COLMAP also adopt skipping way to compute bilateral weighted NCC.}
  \begin{center}
  \begin{tabular}{|c|c|c|c|c|c|}
  \hline
  & $\#$images &  $\eta$ & COLMAP~\cite{Alpher10} & {\bf Ours} & Speed Up \\
  \hline
  \multirow{3}{*}{Fountain} & \multirow{3}{*}{11} & $\times$1 & 506.64 & 177.80 & 2.85 \\
  \cline{3-6}
  & & $\times$0.5 & 202.02 & 44.16 & 4.57 \\
  \cline{3-6}
  & & $\times$0.25 & 88.38 & 12.41 & 7.12 \\
  \hline
  \multirow{3}{*}{Herzjesu} & \multirow{3}{*}{8} & $\times$1 & 363.9 & 92.57 & 3.93 \\
  \cline{3-6}
  & & $\times$0.5 & 142.80 & 23.19 & 6.16 \\
  \cline{3-6}
  & & $\times$0.25 & 63.36 & 7.28 & 8.70 \\
  \hline
  \end{tabular}
  \end{center}
  \label{tab:depthtime}
\end{table}

\section{Conclusion}
In this paper, we present an efficient multi-view stereo method with asymmetric checkerboard propagation and multi-hypothesis joint view selection. The asymmetric checkerboard propagation can propagate the hypotheses with high confidence further and generate more hypotheses. By exploiting the hypotheses generated by the above propagation scheme, the multi-hypothesis joint view selection first constructs a cost matrix at each pixel, and then fully explores the photometric information from the dimension of ``view" and ``hypothesis" to infer the discriminative aggregated view subset parallel. The evaluation on depth map and point cloud shows our method is capable of providing highly accurate and complete reconstruction results with good efficiency.

\bibliographystyle{splncs}
\bibliography{egbib}
\end{document}